\documentclass[10pt,twocolumn,letterpaper]{article}

\usepackage{cvpr}              

\usepackage{graphicx}
\usepackage{amsmath}
\usepackage{amssymb}
\usepackage{booktabs}
\usepackage{multirow}
\usepackage{soul}
\usepackage{booktabs}

\usepackage[normalem]{ulem}
\usepackage[super]{nth}
\useunder{\uline}{\ul}{}
\usepackage[pagebackref,breaklinks,colorlinks]{hyperref}

\usepackage{float}
\usepackage{xcolor}
\usepackage[accsupp]{axessibility}
\usepackage[capitalize]{cleveref}
\crefname{section}{Sec.}{Secs.}
\Crefname{section}{Section}{Sections}
\Crefname{table}{Table}{Tables}
\crefname{table}{Tab.}{Tabs.}

\def\confName{CVPR}
\def\confYear{2022}

\begin{document}

\title{Attentions Help CNNs See Better: Attention-based Hybrid Image Quality Assessment Network}

\author{Shanshan Lao\textsuperscript{1}\thanks{Contribute equally.}\quad Yuan Gong\textsuperscript{1}\footnotemark[1]\quad Shuwei Shi\textsuperscript{1}\quad  Sidi Yang\textsuperscript{1}\quad Tianhe Wu\textsuperscript{1}\quad \\Jiahao Wang\textsuperscript{1}\quad Weihao Xia\textsuperscript{2}\quad Yujiu Yang\textsuperscript{1}\thanks{Corresponding author.}\\
\textsuperscript{1} Tsinghua Shenzhen International Graduate School, Tsinghua University\\
\textsuperscript{2} University College London\\
{\tt\small \{laoss21,\ gang-y21,\ ssw20,\ yangsd21, wang-jh19\}@mails.tsinghua.edu.cn}\\
{\tt\small xiawh3@outlook.com, yang.yujiu@sz.tsinghua.edu.cn}
}
\maketitle


\begin{abstract}
Image quality assessment (IQA) algorithm aims to quantify the human perception of image quality.
Unfortunately, there is a performance drop when assessing the distortion images generated by generative adversarial network (GAN) with seemingly realistic textures. 
In this work, we conjecture that this maladaptation lies in the backbone of IQA models, where patch-level prediction methods use independent image patches as input to calculate their scores separately, but lack spatial relationship modeling among image patches. 
Therefore, we propose an \textbf{A}ttention-based \textbf{H}ybrid \textbf{I}mage \textbf{Q}uality Assessment Network (AHIQ) to deal with the challenge and get better performance on the GAN-based IQA task. 
Firstly, we adopt a two-branch architecture, including a vision transformer (ViT) branch and a convolutional neural network (CNN) branch for feature extraction. 
The hybrid architecture combines interaction information among image patches captured by ViT and local texture details from CNN. 
To make the features from the shallow CNN more focused on the visually salient region, a deformable convolution is applied with the help of semantic information from the ViT branch. 
Finally, we use a patch-wise score prediction module to obtain the final score. 
The experiments show that our model outperforms the state-of-the-art methods on four standard IQA datasets and AHIQ ranked first on the Full Reference (FR) track of the NTIRE 2022 Perceptual Image Quality Assessment Challenge.
Code and pretrained models are publicly available at \url{https://github.com/IIGROUP/AHIQ}
\end{abstract}

\begin{figure}[th]
\centering
\def\svgwidth{\columnwidth}
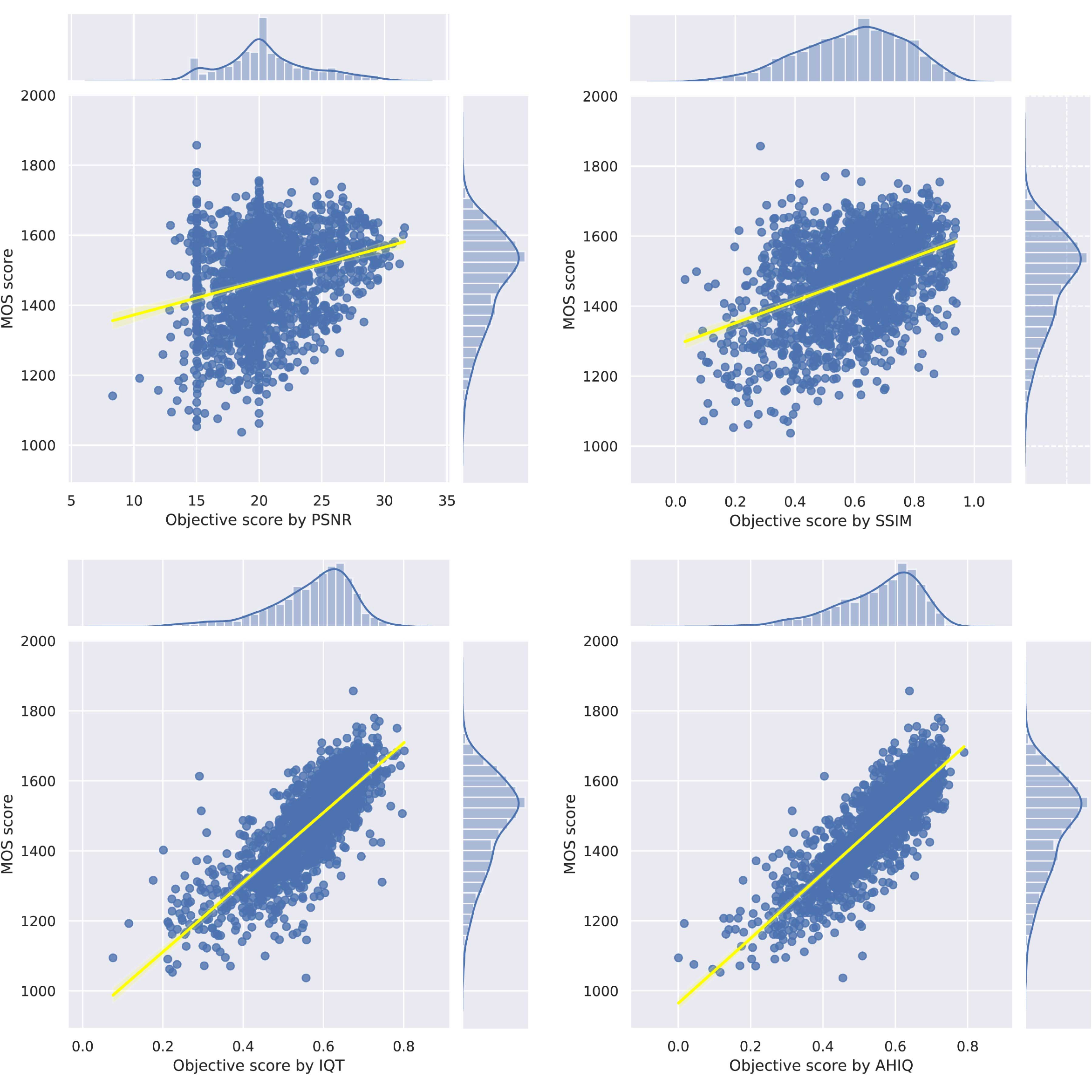
\caption{Scatter plots of the objective scores vs. the MOS scores on the validation dataset of the NTIRE 2022 Perceptual Image Quality Assessment Challenge~\cite{gu2022ntire}. Higher correlation means better performance of the IQA method.}
\label{fig:scatter}
\end{figure}

\section{Introduction}
\label{sec:intro}
Image quality has become a critical evaluation metric in most image-processing applications, including image denoising, image super-resolution, compression artifacts reduction, \etc. Directly acquiring perceptual quality scores from human observers is accurate. However, this requires time-consuming and costly subjective experiments. The goal of Image Quality Assessment (IQA) is to allow computers to simulate the Human Visual System (HVS) through algorithms to score the perceptual quality of images. In this case, the images to be evaluated are often degraded during compression, acquisition, and post-processing. 

In recent years, the invention of Generative Adversarial Networks (GANs)~\cite{goodfellow2014generative} has greatly improved the image processing ability, especially image generation~\cite{gu2020image,xia2021tedigan} and image restoration~\cite{wang2018esrgan}, while it also brings new challenges to image quality assessment. GAN-based methods can fabricate seemingly realistic but fake details and textures~\cite{jinjin2020pipal}. In detail, it is hard for the HVS to distinguish the misalignment of the edges and texture decreases in the region with dense textures. As long as the semantics of textures are similar, the HVS will ignore part of the subtle differences of textures. Most IQA methods for traditional distortion images assess image quality through pixel-wise comparison, which will lead to underestimation for GAN-generated images~\cite{wang2004image}. To deal with the texture misalignment, recent studies~\cite{bosse2017deep} introduce patch-wise prediction methods. 
Some following studies~\cite{shi2021region,jinjin2020pipal} further propose different spatially robust comparison operations into the CNN-based IQA network. However, they take each patch as an independent input and separately calculate their score and weight, which will lead to the loss of context information and the inability to model the relationship between patches. 

Therefore, on the basis of patch-level comparison, we need to better model the interrelationship between patches. To this end, we use Vision Transformer (ViT)~\cite{dosovitskiy2020image} as a feature extractor, which can effectively capture long-range dependencies among patches through a multi-head attention mechanism. However, the vanilla ViT uses a large convolution kernel to down-sample the input images in spatial dimension before entering the network; some details that should be considered are lost, which are also crucial to image quality assessment. Based on the observation, we found that a shallow CNN is a good choice to provide detailed spatial information. 
The features extracted by a shallow CNN contains unwanted noises and merging ViT features with them would decrease the performance. To alleviate the impact of noise, we propose to mimic the characteristic of the HVS that human always pay attention to the salient regions of images. 
Instead of injecting the complete features from a shallow CNN into those from ViT, we only use those that convey spatial details of the salient regions for image quality assessment, thereby alleviating the aforementioned noise.
Furthermore, using max-pooling or average-pooling to directly predict the score of an image will lose crucial information.
Therefore, we use an adaptive weighted strategy to predict the score of an image.


In this work, we introduce an effective hybrid architecture for image quality assessment, which leverages local details from a shallow CNN and global semantic information captured by ViT to further improve IQA accuracy. Specifically, we first adopt a two-branch feature extractor. Then, we use semantic information captured by ViT to find the salient region in images through deformable convolution~\cite{dai2017deformable}. Based on the consideration that each pixel in the deep feature map corresponds to different patches of the input image, we introduce the patch-wise prediction module, which contains two branches, one to calculate a score for each image patch, the other one to calculate the weight of each score.

Extensive experiments show that our method outperforms current approaches in four benchmark image quality assessment datasets~\cite{sheikh2006statistical,larson2010most,ponomarenko2015image,jinjin2020pipal}. The scatter diagram of the correlation between predicted scores and MOS is shown in \cref{fig:scatter} where the plot for IQT is from our own implementation. Visualization experiments reveal that the proposed method is almost linear with MOS, which means that we can better imitate human image perception. Our primary contributions can be summarized as follows:
\begin{itemize}
\item We propose an effective hybrid architecture for image quality assessment, which compares images at the patch level, adds spatial details as a supplement, and scores images patch by patch, considering the relationship between patches and different contributions from each patch.

\item Our method outperforms the state-of-the-art approaches on four benchmark image quality assessment datasets. In particular, the proposed architecture achieves outstanding performance on the PIPAL dataset with various GAN-based distortion and ranked first in the NTIRE 2022 challenge on perceptual image quality assessment.
\end{itemize}

\section{Related Works}
\label{sec:related}
\subsection{Image Quality Assessment}
The goal of IQA is to mimic the HVS to rate the perceived quality of an image accurately. Although it's easy for human beings to assess an image's perceptual quality, IQA is considered to be difficult for machines. Depending on the scenarios and conditions, current IQA methods can be divided into three categories: full-reference (FR) ,and no-reference (NR) IQA. FR-IQA methods take the distortion image and the corresponding reference image as inputs to measure their perceptual similarity. The most widely-used FR-IQA metrics are PSNR and SSIM~\cite{wang2004image} which are conventional and easy to optimize.
\begin{figure*}[th]
  \centering
   \includegraphics[scale=0.48]{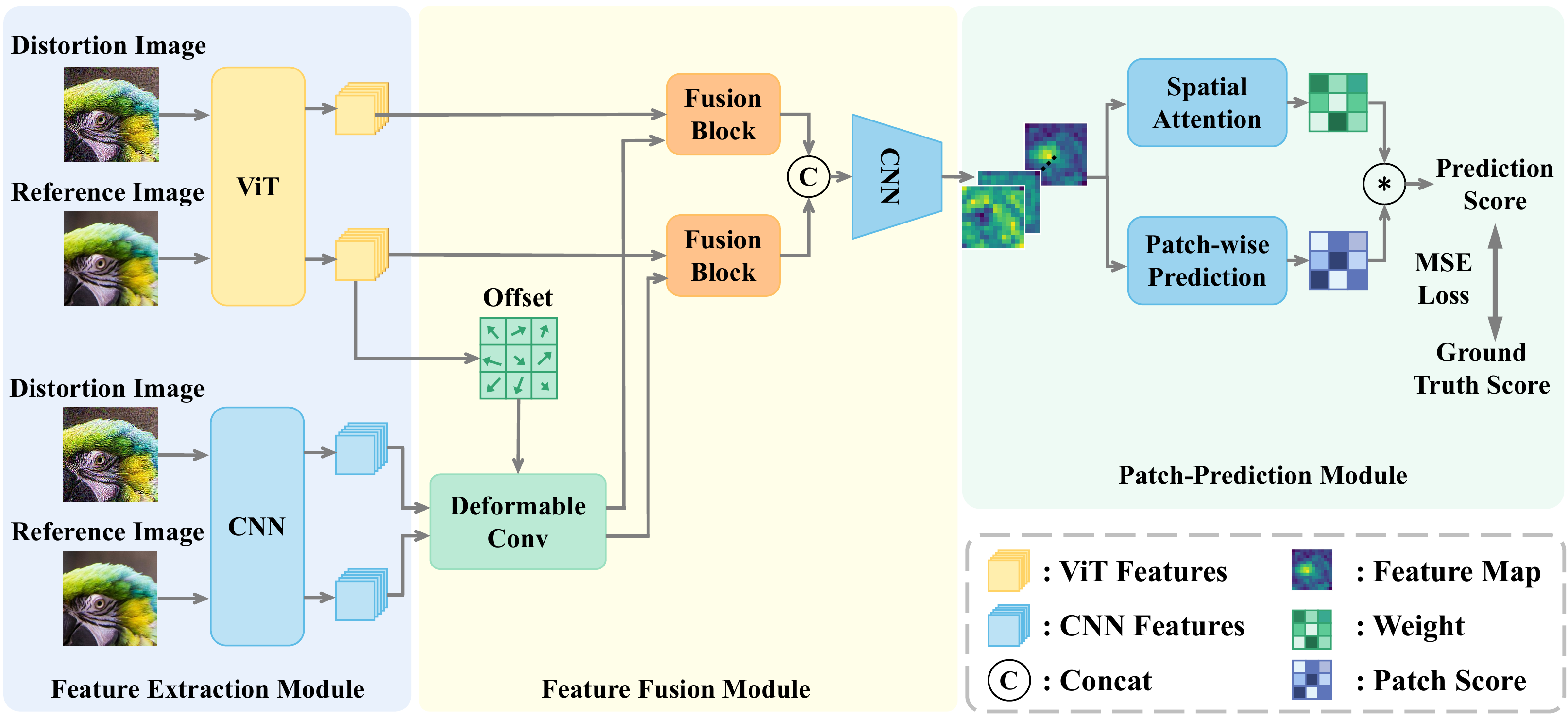}
   \caption{Overview of AHIQ. The proposed model takes a pair of the reference image and distortion image as input and then obtains feature maps through ViT~\cite{dosovitskiy2020image} and CNN, respectively. The feature maps of reference image from ViT are used as global information to obtain the offset map of the deformable convolution\cite{dai2017deformable}. After the feature fusion module which fuses the feature maps, we use a patch-wise prediction module to predict a score for each image patch. The final output is the weighted sum of the scores.}
   \label{fig1:arch}
\end{figure*}
Apart from the conventional IQA methods, various learning-based FR-IQA methods~\cite{zhang2018unreasonable,bosse2017deep,prashnani2018pieapp} have been proposed to address the limitations of conventional IQA methods recently. Zhang \etal~\cite{zhang2018unreasonable} proposed to use the learned perceptual image patch similarity (LPIPS) metric for FR-IQA and proved that deep features obtained through pre-trained DNNs outperform previous classic metrics by large margins. WaDIQaM~\cite{bosse2017deep} is a general end-to-end deep neural network that enables jointly learning of local quality and local weights. PieAPP~\cite{prashnani2018pieapp} is proposed to learn to rank rather than learn to score, which means the network learns the probability of preference of one image over another. IQT~\cite{cheon2021perceptual} applies an encoder-decoder transformer architecture with trainable extra quality embedding and ranked first place in NTIRE 2021 perceptual image quality assessment challenge. In addition, common CNN-based NR-IQA methods~\cite{su2020blindly,wu2020end,xia2020domain} directly extract features from the low-quality images and outperform traditional handcrafted approaches. You \etal~\cite{ you2021transformer} introduced transformer architecture for the NR-IQA recently.


\subsection{Vision Transformer}
Transformer architecture based on self-attention mechanism~\cite{vaswani2017attention} was first proposed in the field of Natural Language Processing (NLP) and significantly improved the performances of many NLP tasks thanks to its representation capability. Inspired by its success in NLP, efforts are made to apply transformers to vision tasks such as image classification~\cite{dosovitskiy2020image}, object detection~\cite{carion2020end, zhu2020deformable}, low-level vision~\cite{yang2020learning}, \etc. Vision Transformer (ViT) introduced by Dosovitskiy~\etal~\cite{dosovitskiy2020image} is directly inherited from NLP, but takes raw image patches as input instead of word sequences. ViT and its follow-up studies have become one of the mainstream feature extraction backbones except for CNNs.

Compared with the most commonly used CNNs, transformer can derive global information while CNNs mainly focus on local features. In IQA tasks, global and local information are both crucial to the performance because when human beings assess image quality, both the information are naturally taken into account. Inspired by this assumption, we propose to combine long-distance features and local features captured by ViT and CNNs, respectively. To fulfill this goal, we use a two-branch feature extraction backbone and feature fusion modules, which will be detailed in~\cref{sec:method}.

\subsection{Deformable Convolution}
Deformable convolution~\cite{dai2017deformable} is an efficient and powerful mechanism which is first proposed to deal with sparse spatial locations in high-level vision tasks such as object detection~\cite{dai2017deformable,bertasius2018object, zhu2019deformable }, semantic segmentation~\cite{zhu2019deformable}, and human pose estimation~\cite{sun2018integral}. By using deformed sampling locations with learnable offsets, deformable convolution enhances the spatial sampling locations and improves the transformation modeling ability of CNNs. Recently, deformable convolution continues its strong performance in low-level vision tasks including video deblurring~\cite{wang2019edvr}, video super-resolution~\cite{chan2021understanding}. It is first combined with IQA methods by Shi~\etal~\cite{shi2021region} to perform a reference-oriented deformable convolution in the full-reference scenario.

\section{Methodology}
\label{sec:method}

In this section, we introduce the overall framework of the Attention-based Hybrid Image Quality Assessment Network (AHIQ). As shown in Fig \ref{fig1:arch}, the proposed network takes pairs of reference images and distortion images as input, and it consists of three key components: a feature extraction module, a feature fusion module, and a patch-wise prediction module. 

For the reason that GAN-based image restoration methods~\cite{wang2018esrgan,gu2020image} often fabricate plausible details and textures, it is difficult for the network to distinguish GAN-generated texture from noise and real texture by pixel-wise image difference. Our proposed model aims to deal with it. We employ the Vision Transformer to model the relationship and capture long-range dependencies among patches. Shallow CNN features are introduced to add detailed spatial information. In order to help CNN focus on the salient region, we use deformable convolution guided by semantic information from ViT. We use an adaptive weighted scoring mechanism to give a comprehensive assessment.

\subsection{Feature Extraction Module}
\label{subsec:extraction}
\begin{figure}[th]
\centering
\includegraphics[scale=0.41]{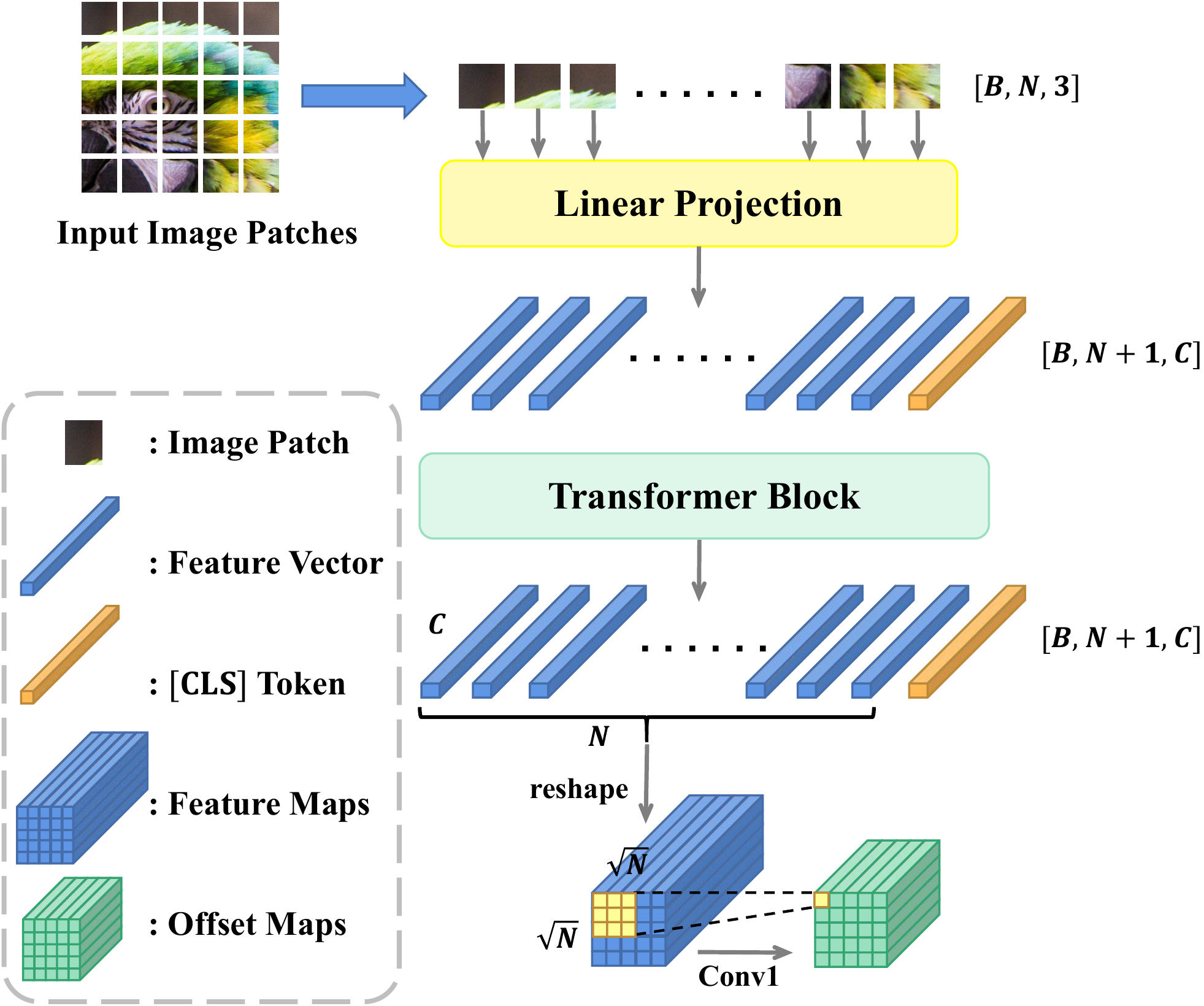}
\caption{The illustration of vision Transformer for feature extraction module. The class token (orange) is regarded when the feature maps are extracted.
}
\label{fig2:feat}
\end{figure}
As is depicted in \cref{fig1:arch}, the front part of the architecture is a two-branch feature extraction module that consists of a ViT branch and a CNN branch. The transformer feature extractor mainly focuses on extracting global and semantic representations. Self-attention modules in transformer enable the network to model long-distance features and encode the input image patches into feature representations. Patch-wise encoding is helpful to assess the output image quality of GAN-based image restoration because it enhances the tolerance of spatial misalignment. Since humans also pay attention to details when judging the quality of an image, so detailed and local information is also important. To this end, we introduce another CNN extraction branch apart from the transformer branch to add more local textures.

In the forward process, a pair of the reference image and distortion image are fed into the two branches, respectively, and we then take out their feature maps in the early stages. For the transformer branch, as illustrated in~\cref{fig2:feat}, output sequences from Vision Transformer~\cite{dosovitskiy2020image} are reshaped into feature maps $f_{T}\in \mathbb{R}^{p\times p \times 5c}$ discarding the class token, where $p$ represent the size of the feature map. For the CNN branch, we extract shallow feature map from ResNet~\cite{he2016deep} $f_{C}\in \mathbb{R}^{4p\times 4p \times C}$ where $C=256\times 3$. Finally, we put the obtained feature maps into the feature fusion module, which will be specified next.

\subsection{Feature Fusion Module}
\label{subsec:fusion}
We argue that feature maps from the early stages of CNN provide low-level texture details but bring along some noise. To address this problem, we take advantage of transformer architecture to capture global and semantic information. In our proposed network, feature maps from ViT with rich semantic information are used to find the salient region of the image. This perception procedure is performed in a content-aware manner and allows the network better mimic the way humans perceive image quality. Particularly, the feature maps from ViT are used to learn an offset map for deformable convolution as is shown in~\cref{fig2:feat}. Then we perform this deformable convolution~\cite{dai2017deformable} operation on feature maps from CNN, which we elaborate on previously. In this way, features from a shallow CNN can be better modified and utilized for further feature fusion. Obviously, in the previous description, feature maps from the two branches differ from each other in spatial dimension and need to be aligned. Therefore, a simple 2-layer convolution network is applied to project the feature maps after deformable convolution to the same width~\textit{W} and height~\textit{H} with ViT. The whole process can be formulated as follows:
\begin{align}
    && \Delta p &= \text{Conv1}(f_{T})), \\
    &&  f_{C} &= \text{DConv}(f_{org},\Delta p), \\
    &&  f_{C}^{'} &= \text{Conv2}(\text{ReLU}(\text{Conv2}(f_{C}))),   \\
    &&  f^{u} &= \text{Concat}[f_{T},f_{C}^{'}],   \\
    &&  f_{all} &= \text{Concat}[f^{u}_{dis},f^{u}_{ref},f^{u}_{dis}-f^{u}_{ref}],  \\
    &&  f_{out} &= \text{Conv3}(\text{ReLU}(\text{Conv3}(f_{all}))),
\end{align}
where $f_T$ denotes feature maps from the transformer branch, $\Delta p$ denotes offset map, $f_{org}$ and $f_C$ denote feature maps from CNN, DConv means deformable convolution. Note that Conv2 is a convolution operation with a stride of 2, downsampling $f_C \in \mathbb{R}^{4p \times 4p \times C}$ by four times to $f_C^{'} \in \mathbb{R}^{p \times p \times C}$. 

\subsection{Patch-wise Prediction Module}
\label{subsec:pooling}
\begin{figure}[th]
\centering
\includegraphics[scale=0.43]{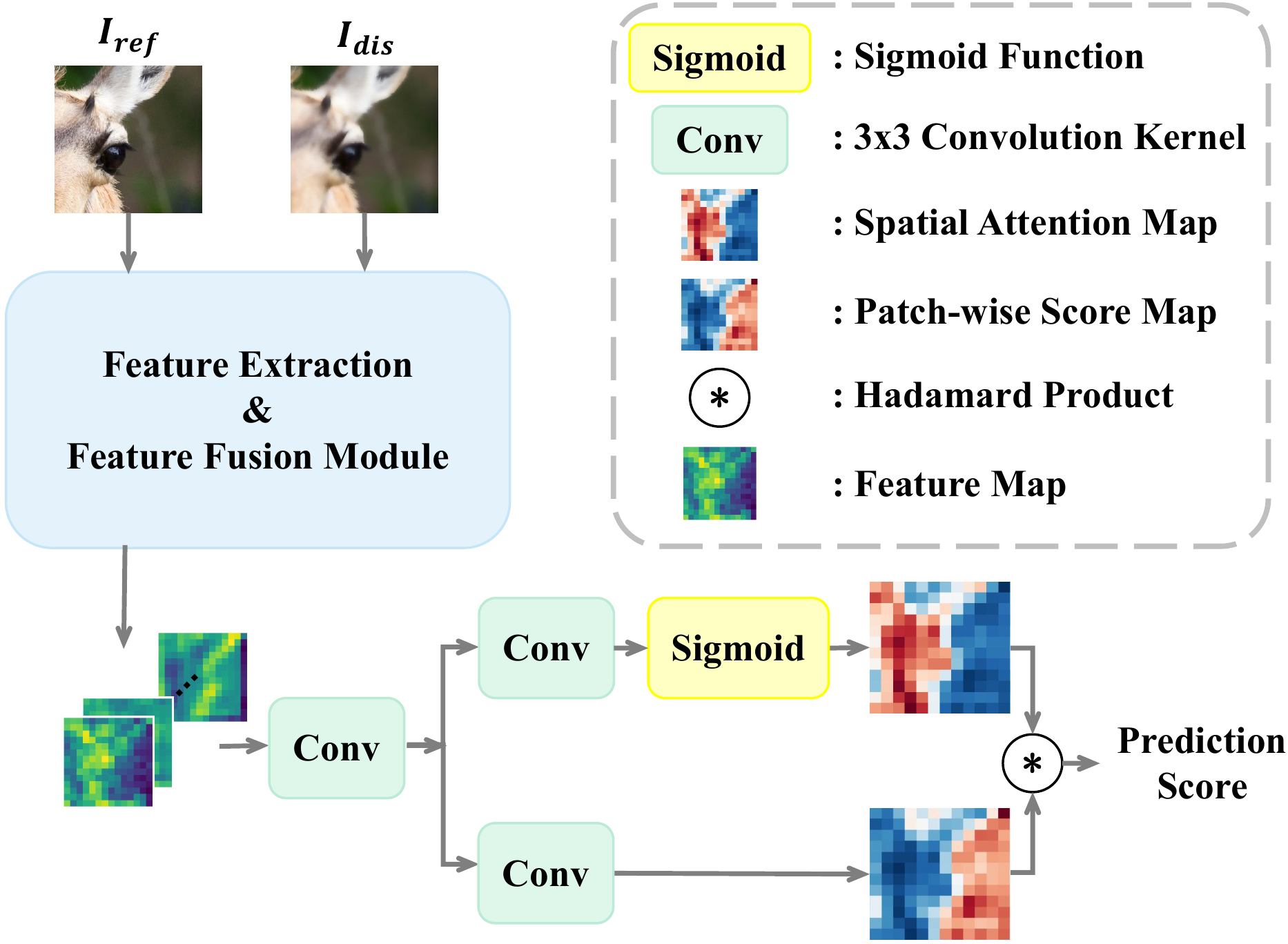}
\caption{The pipeline of the proposed patch-wise prediction module. This two-branch module takes feature maps as input, then generates a patch-wise score map and its corresponding attention map to obtain the final prediction by weighted average.}
\label{fig:pixel}
\end{figure}

Given that each pixel in the deep feature map corresponds to a different patch of the input image and contains abundant information, the information in the spatial dimension is indispensable. However, in previous works, spatial pooling methods such as max-pooling and average-pooling are applied to obtain a final single quality score. This pooling strategy loses some information and ignores the relationships between image patches. Therefore, we introduce a two-branch patch-wise prediction module which is made up of a prediction branch and a spatial attention branch, as illustrated in \cref{fig:pixel}. The prediction branch calculates a score for each pixel in the feature map, while the spatial attention branch calculates an attention map for each corresponding score. Finally, we can obtain the final score by weighted summation of scores. The weighted sum operation helps to model the significance of the region to simulate the human visual system. This can be expressed as follows:
\begin{equation}
    s_f = \frac{\textbf{s} * \textbf{w}}{\sum \textbf{w}},
\end{equation}
where $\textbf{s}\in \mathbb{R}^{H\times W \times 1}$ denotes score map, $\textbf{w}\in \mathbb{R}^{H\times W \times 1}$ denotes the corresponding attention map, $*$ means Hadamard product and $s_f$ means the final predicted score. MSE loss between the predicted score and the ground truth score is utilized for the training process in our proposed method.

\begin{table*}[th]
\centering
\renewcommand{\arraystretch}{1.08}
\caption{IQA datasets for performance evaluation and model training.}
\begin{tabular}{cccccccc}
\toprule[1.2pt]
Database  & \# Ref & \# Dist & Dist. Type & \# Dist. Type & Rating & Rating Type & Env. \\ \hline
LIVE~\cite{sheikh2006statistical}      & 29    & 779    & traditional       & 5            & 25k    & MOS         & lab           \\
CSIQ~\cite{larson2010most}      & 30    & 866    & traditional       & 6            & 5k     & MOS         & lab           \\
TID2013~\cite{ponomarenko2015image}   & 25    & 3,000  & traditional       & 25           & 524k   & MOS         & lab           \\
KADID-10k~\cite{lin2019kadid} & 81    & 10.1k  & traditional       & 25           & 30.4k  & MOS         & crowdsourcing \\
PIPAL~\cite{jinjin2020pipal}     & 250   & 29k    & trad.+alg.outputs & 40           & 1.13m  & MOS         & crowdsourcing \\ \toprule[1.2pt]
\label{tab:data}
\end{tabular}
\end{table*}

\begin{table*}[th]
\centering
\caption{Performance comparisons on LIVE, CSIQ, and TID2013 Databases. Performance scores of other methods are as reported in the corresponding original papers and~\cite{ding2020image}. The best scores are~\textbf{bolded} and missing scores are shown as ``–'' dash.}
\setlength{\tabcolsep}{6mm}{
\begin{tabular}{ccccccc}
\toprule[1.2pt]
\multirow{2}{*}{Method} & \multicolumn{2}{c}{LIVE} & \multicolumn{2}{c}{CSIQ} & \multicolumn{2}{c}{TID2013} \\ \cline{2-7} 
                        & PLCC        & SROCC      & PLCC        & SROCC      & PLCC         & SROCC        \\ \hline
PSNR                    & 0.865       & 0.873      & 0.819       & 0.810      & 0.677       & 0.687        \\
SSIM~\cite{wang2004image}                    & 0.937       & 0.948      & 0.852       & 0.865      & 0.777        & 0.727        \\
MS-SSIM~\cite{wang2003multiscale}                 & 0.940       & 0.951      & 0.889       & 0.906      & 0.830        & 0.786        \\
FSIMc~\cite{zhang2011fsim}                   & 0.961       & 0.965      & 0.919       & 0.931      & 0.877        & 0.851        \\
VSI~\cite{zhang2014vsi}                     & 0.948       & 0.952      & 0.928       & 0.942      & 0.900          & 0.897        \\
MAD~\cite{larson2010most} &0.968      &0.967      &0.950       &0.947     &0.827      &0.781      \\
VIF~\cite{sheikh2006image} &0.960      &0.964      &0.913      &0.911      &0.771      &0.677      \\
NLPD~\cite{laparra2016perceptual}    &0.932      &0.937      &0.923      &0.932      &0.839      &0.800      \\
GMSD~\cite{xue2013gradient}    &0.957      &0.960      &0.945      &0.950      &0.855      &0.804\\    
SCQI~\cite{bae2016novel}                    & 0.937       & 0.948      & 0.927       & 0.943      & 0.907        & 0.905        \\ \hline
DOG-SSIMc~\cite{pei2015image}               & 0.966       & 0.963      & 0.943       & 0.954      & 0.934        & 0.926        \\
DeepQA~\cite{kim2017deep}                  & 0.982       & 0.981      & 0.965       & 0.961      & 0.947        & 0.939        \\
DualCNN~\cite{varga2020composition}                 & -           & -          & -           & -          & 0.924        & 0.926        \\
WaDIQaM-FR~\cite{bosse2017deep}              & 0.98        & 0.97       & -           & -          & 0.946        & 0.94         \\
PieAPP~\cite{prashnani2018pieapp}                  & 0.986       & 0.977      & 0.975       & 0.973      & 0.946        & 0.945        \\
JND-SalCAR~\cite{seo2020novel} & 0.987          & \textbf{0.984} & 0.977          & \textbf{0.976} & 0.956          & 0.949          \\
AHIQ (ours) & \textbf{0.989} & \textbf{0.984} & \textbf{0.978} & 0.975          & \textbf{0.968} & \textbf{0.962}       \\ \toprule[1.2pt]

\end{tabular}}
\label{tab:sota}
\end{table*}

\section{Experiment}
\subsection{Datasets}

We employ four datasets that are commonly used in the research of perceptual image quality assessment, including LIVE~\cite{sheikh2006statistical}, CSIQ~\cite{larson2010most}, TID2013~\cite{ponomarenko2015image}, and PIPAL~\cite{jinjin2020pipal}. \cref{tab:data} compares the listed datasets in more detail. In addition to PIPAL, the other datasets only include traditional distortion types, while PIPAL includes a large number of distorted images including GAN-generated images. 

As recommended, we randomly split the datasets into training (60\%), validation (20\%), and test set (20\%) according to reference images. Therefore, the test data and validation data will not be seen during the training procedure. We use the validation set to select the model with the best performance and use the test set to evaluate the final performance.

\subsection{Implementation Details}
Since we use ViT~\cite{dosovitskiy2020image} and ResNet~\cite{he2016deep} models pre-trained on ImageNet~\cite{ILSVRC15}, we normalize all input images and randomly crop them into $224\times224$. We use the outputs of five intermediate blocks $\{0,1,2,3,4\}$ in ViT, each of which consists of a self-attention module and a Feed-Forward Network (FFN). The feature map from one block $f\in \mathbb{R}^{p\times p \times c}$, where $c=768, p=14 \ \text{or} \ 28$, are concatenated into $f_{T}\in \mathbb{R}^{p\times p \times 6c}$. We also take out the output feature maps from all the 3 layers in stage 1 of ResNet and concatenate them together to get $f_{C}\in \mathbb{R}^{56\times 56 \times C}$ where $C=256\times 3$. And random horizontal flip rotation is applied during the training. The training loss is computed using a mean squared error (MSE) loss function. During the validation phase and test phase, we randomly crop each image 20 times and the final score is the average score of each cropped image. It should be noted that we use pretrained ViT-B/16 as the backbone in all experiments on traditional datasets including LIVE, CSIQ and TID2013, while ViT-B/8 is utilized in PIPAL.

For optimization, we use the AdamW optimizer with an initial learning rate $lr$ of $10^{-4}$ and weight decay of $10^{-5}$. We set the minibatch size as 8. Set the learning rate of each parameter group using a cosine annealing schedule, where $\eta_{max}$ is set to the initial $lr$ and the number of epochs $T_{cur}$ is 50. 
We implemented our proposed model AHIQ in Pytorch and trained using a single NVIDIA GeForce RTX2080 Ti GPU. The practical training runtimes differ across datasets as the number of images in each dataset is different. Training one epoch on the PIPAL dataset requires thirty minutes.

\subsection{Comparison with the State-of-the-art Methods}
We assess the performance of our model with Pearson’s linear
correlation coefficient (PLCC) and Spearman’s rank-order correlation coefficient (SROCC). 
PLCC assesses the linear correlation between ground truth and the predicted quality scores, whereas SROCC describes the level of monotonic correlation.

\vspace{2pt}
\noindent\textbf{Evaluation on Traditional Dataset.} We evaluate the effectiveness of AHIQ on four benchmark datasets. For all our tests, we follow the above experimental setup. It can be shown in~\cref{tab:sota} that AHIQ outperforms or is competitive with WaDIQaM~\cite{bosse2017deep}, PieAPP~\cite{prashnani2018pieapp}, and JND-SalCAR~\cite{seo2020novel} for all tested datasets. Especially on the more complex dataset TID2013, our proposed model achieved a solid improvement over previous work. This shows that the AHIQ can cope well with different types of distorted images.
\begin{table}[th]
\centering
\caption{Performance comparison after training on the entire KADID dataset~\cite{lin2019kadid}, then test on LIVE, CSIQ, and TID2013 Databases. Part of the performance scores of other methods are borrowed from~\cite{ding2020image}. The best scores are~\textbf{bolded} and missing scores are shown as ``–'' dash.}
\scalebox{0.82}{
\begin{tabular}{cccc}
\toprule[1.2pt]
\multirow{2}{*}{Method} & LIVE                 & CSIQ                 & TID2013              \\ \cline{2-4} 
& PLCC/SROCC           & PLCC/SROCC           & PLCC/SROCC           \\ \hline

WaDIQaM~\cite{bosse2017deep}              & 0.940/0.947       & 0.901/0.909          & 0.834/0.831         \\
PieAPP~\cite{prashnani2018pieapp}                  & 0.908/0.919      & 0.877/0.892      & 0.859/0.876        \\
LPIPS~\cite{zhang2018unreasonable}              & 0.934/0.932  &0.896/0.876  &0.749/0.670           \\
DISTS~\cite{ding2020image}               & \textbf{0.954}/0.954        & 0.928/0.929      & 0.855/0.830     \\
IQT~\cite{cheon2021perceptual}                 & -/\textbf{0.970}      & -/0.943      & -/0.899      \\
AHIQ (ours) & 0.952/\textbf{0.970} & \textbf{0.955/0.951}        & \textbf{0.899/0.901}       \\ \toprule[1.2pt]
\end{tabular}
}
\label{tab:kadid}
\end{table}

\vspace{2pt}
\noindent\textbf{Evaluation on PIPAL.} We compare our models with the state-of-the-art FR-IQA methods on the NTIRE 2022 IQA challenge validation and testing datasets. As shown in \cref{tab:pipal}, AHIQ achieves outstanding performance in terms of PLCC and SROCC compared with all previous work. In particular, our method substantially outperforms IQT, which is recognized as the first transformer-based image quality assessment network, through the effective feature fusion from the shallow CNN and ViT as well as the proposed patch-wise prediction module. This verifies the effectiveness of our model for GAN-based distortion image quality assessment.

\begin{table}[th]
\centering
\caption{
Performance comparison of different IQA methods on PIPAL dataset. AHIQ-C is the ensemble version we used for the NTIRE 2022 Perceptual IQA Challenge.
}
\begin{tabular}{ccccc}
\toprule[1.2pt]
\multirow{2}{*}{Method} & \multicolumn{2}{c}{Validation} & \multicolumn{2}{c}{Test} \\ \cline{2-5} 
 & PLCC & SROCC & PLCC & SROCC \\ \hline
PSNR & 0.269 & 0.234 & 0.277 & 0.249 \\
NQM~\cite{damera2000image} & 0.364 & 0.302 & 0.395 & 0.364 \\
UQI~\cite{wang2002universal} & 0.505 & 0.461 & 0.450 & 0.420 \\
SSIM~\cite{wang2004image} & 0.377 & 0.319 & 0.391 & 0.361 \\
MS-SSIM~\cite{wang2003multiscale} & 0.119 & 0.338 & 0.163 & 0.369 \\
RFSIM~\cite{zhang2010rfsim} & 0.285 & 0.254 & 0.328 & 0.304 \\
GSM~\cite{liu2011image} & 0.450 & 0.379 & 0.465 & 0.409 \\
SRSIM~\cite{zhang2012sr} & 0.626 & 0.529 & 0.636 & 0.573 \\
FSIM~\cite{zhang2011fsim} & 0.553 & 0.452 & 0.571 & 0.504 \\
VSI~\cite{zhang2014vsi} & 0.493 & 0.411 & 0.517 & 0.458 \\
NIQE~\cite{mittal2012making} & 0.129 & 0.012 & 0.132 & 0.034 \\
MA~\cite{ma2017learning} & 0.097 & 0.099 & 0.147 & 0.140 \\
PI~\cite{blau2018perception} & 0.134 & 0.064 & 0.145 & 0.104 \\
Brisque~\cite{mittal2011blind} & 0.052 & 0.008 & 0.069 & 0.071 \\ \hline
LPIPS-Alex~\cite{zhang2018unreasonable} & 0.606 & 0.569 & 0.571 & 0.566 \\
LPIPS-VGG~\cite{zhang2018unreasonable} & 0.611 & 0.551 & 0.633 & 0.595 \\
DISTS~\cite{ding2020image} & 0.634 & 0.608 & 0.687 & 0.655 \\
IQT~\cite{cheon2021perceptual} & 0.840 & 0.820 & 0.799 & 0.790 \\ \hline
AHIQ (ours) & 0.845 & 0.835 & 0.823 & 0.813 \\
AHIQ-C (ours) & \textbf{0.865} & \textbf{0.852} & \textbf{0.828} & \textbf{0.822}  \\ \toprule[1.2pt]
\end{tabular}
\label{tab:pipal}
\end{table}

\vspace{2pt}
\noindent\textbf{Cross-Database Performance Evaluation.} 
To evaluate the generalization of our proposed AHIQ, we conduct the cross-dataset evaluation on LIVE, CSIQ, and TID2013. We train the model on KADID and the training set of PIPAL respectively. Then we test it on the full set of the other three benchmark datasets. As shown in \cref{tab:kadid} and \cref{tab:cross}, AHIQ achieves satisfactory generalization ability. 

\begin{table}[th]
\centering
\caption{Performance comparison for cross-database evaluations.}
\scalebox{0.82}{
\begin{tabular}{cccc}
\toprule[1.2pt]
\multirow{2}{*}{Method} & LIVE                 & CSIQ                 & TID2013              \\ \cline{2-4} 
                        & PLCC/SROCC           & PLCC/SROCC           & PLCC/SROCC           \\ \hline
PSNR                    & 0.865/0.873          & 0.786/0.809          & 0.677/0.687          \\
WaDIQaM~\cite{bosse2017deep}                 & 0.837/0.883          & -/-                  & 0.741/0.698          \\
RADN~\cite{shi2021region}                    & 0.878/0.905          & -/-                  & 0.796/0.747 \\
AHIQ (ours)              & \textbf{0.911/0.920} & \textbf{0.861/0.865} & \textbf{0.804/0.763} \\ \toprule[1.2pt]
\end{tabular}
}
\label{tab:cross}
\end{table}

\subsection{Ablation Study}
In this section, we analyze the effectiveness of the proposed network by conducting ablation studies on the NTIRE 2022 IQA Challenge testing datasets~\cite{gu2022ntire}. With different configuration and implementation strategies, we evaluate the effect of each of the three major components: feature extraction module, feature fusion module, and patch-wise prediction module.
\begin{table}[ht]
\centering
\caption{Comparison of different feature fusion strategies on the NTIRE 2022 IQA Challenge testing datasets. CNN refers to Resnet50 and ViT refers to ViT-B/8 in this experiment. }

\begin{tabular}{cccccc}
\toprule[1.2pt]
\multirow{2}{*}{No.} & \multicolumn{2}{c}{Feature} & \multirow{2}{*}{Fusion Method} & \multirow{2}{*}{PLCC} & \multirow{2}{*}{SROCC}  \\ \cline{2-3}
 & CNN & ViT &  &  & \\ \hline
1 & \checkmark & \checkmark & deform+concat & \textbf{0.823} & \textbf{0.813}  \\
2 & \checkmark & \checkmark & concat & 0.810 & 0.799 \\
3 & \checkmark &  & - & 0.792 & 0.789 \\
4 &  & \checkmark & - & 0.799 & 0.788 \\ \toprule[1.2pt]
\end{tabular}
\label{tab:fusion}
\end{table}

\vspace{2pt}
\noindent\textbf{Feature Extraction Backbone.} We experiment with different representative feature-extraction backbones and the comparison result is provided in \cref{tab:backbone}. The CNN backbones used for comparison include ResNet50, ResNet101, ResNet152~\cite{he2016deep}, HRNet~\cite{wang2020deep}, Inception-ResNet-V2~\cite{szegedy2017inception}, and the transformer backbones include ViT-B/16 and ViT-B/8~\cite{dosovitskiy2020image}. It is noteworthy that ViT-B consists of 12 transformer blocks and the sizes of the image patches are $16\times 16$ and $8\times 8$ respectively with an input shape of $224\times 224$. 

It can be found that the network using ResNet50 and ViT-B/8 ends up performing the best. The experimental results demonstrate that deeper and wider CNN is unnecessary for AHIQ. We believe this is because CNN plays the role of providing shallow and local feature information in AHIQ. We only take out the intermediate layers from the first stage, so shallow features will contain less information when the network is too deep or too complicated.

\begin{table}[ht]
\centering
\caption{Comparison of different feature extraction backbones on the NTIRE 2022 IQA Challenge testing datasets.}
\scalebox{0.95}{
\begin{tabular}{ccccc}
\toprule[1.2pt]
CNN & ViT & PLCC & SROCC & Main Score \\ \hline
Resnet50 & \multirow{5}{*}{ViT-B/8} & \textbf{0.823} & \textbf{0.813} & \textbf{1.636} \\
Resnet101 &  & 0.802 & 0.788 & 1.590 \\
Resnet152 &  & 0.807 & 0.793 & 1.600 \\
HRnet &  & 0.806 & 0.796 & 1.601 \\
IncepResV2 &  & 0.806 & 0.793 & 1.599 \\ \hline
Resnet50 & ViT-B/16 & 0.811 & 0.803 & 1.614 \\ \toprule[1.2pt]
\end{tabular}}
\label{tab:backbone}
\end{table}

\vspace{2pt}
\noindent\textbf{Fusion strategy.} We further examine the effect of features from CNN and ViT as well as the feature fusion strategies. As is tabulated in \cref{tab:fusion}, the first two experiments adopt different methods for feature fusion. The first one is the method we adopt in our AHIQ. For the second experiment, the features from transformer and ViT are simply concatenated together. The first method outperforms the second one by a large margin which demonstrates that using deformable convolution to modify CNN feature maps is well-effective. This further illustrates the power of global and semantic information in transformer to guide the shallow features by paying more attention to the salient regions.

We also conduct ablation studies on using features from ViT and from CNN separately. Results are at the last two rows in~\cref{tab:fusion}. One can observe that only using one of the CNN and Transformer branches results in a dramatic decrease in performance. This experimental result shows that both global semantic information brought by ViT and local texture information introduced by CNN is very crucial in this task, which is well consistent with our previous claim.
\begin{figure}[th]
\centering
\includegraphics[scale=0.26]{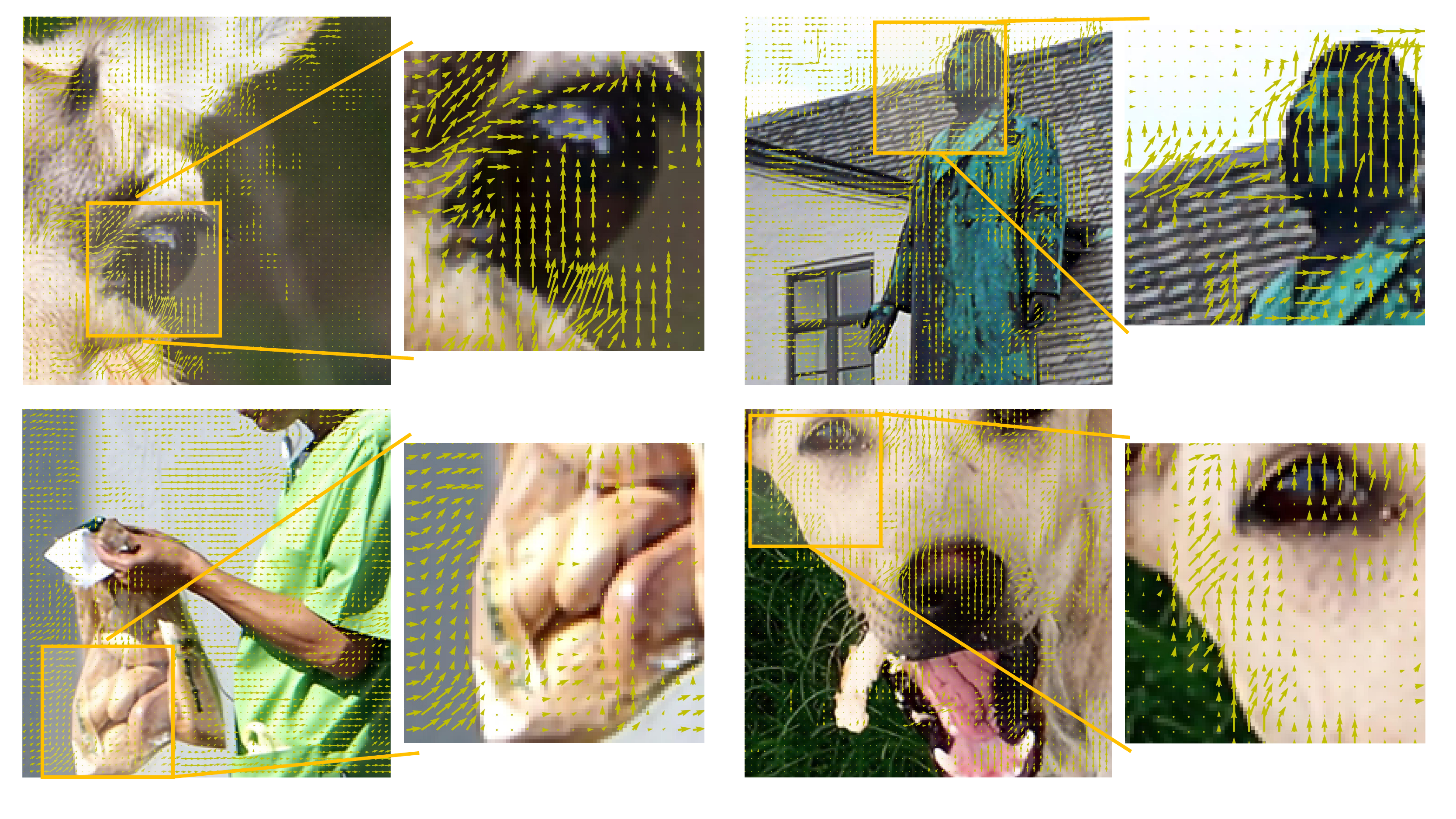}
\caption{The visualization of learned offsets from deformable convolution. For each case, the vector flow which displays the learned offsets and zoomed-in details are included.}
\label{fig:optical}
\end{figure}

\vspace{2pt}
\noindent\textbf{Visualization of Learned Offset.} We visualize the learned offsets from deformable convolution in \cref{fig:optical}. It can be observed that the learned offsets indicated by arrows mainly affect edges and salient regions. In addition, most of the offset vectors point from the background to the salient regions, which means that in the process of convolution, the sampling locations moves to the significant region by the learned offsets. This visualization results illustrate the argument we made earlier that semantic information from ViT help CNN see better by deformable convolution.

\begin{table}[ht]
\centering
\caption{Comparison of different pooling strategy on the NTIRE 2022 IQA Challenge testing datasets. Note that ``Patch'' denotes the patch-wise prediction and ``Spatial'' denotes the spatial pooling.}
\begin{tabular}{cccc}
\toprule[1.2pt]
Pooling Strategy & PLCC & SROCC & Main Score \\ \hline
Patch & \textbf{0.823} & \textbf{0.813} & \textbf{1.636} \\
Spatial & 0.794 & 0.795 & 1.589 \\
Patch + Spatial & 0.801 & 0.791 &  1.593\\ \toprule[1.2pt]
\end{tabular}
\label{tab:pooling}
\end{table}

\vspace{2pt}
\noindent\textbf{Pooling Strategy.} Experiments on different pooling strategies are conducted, and the results are shown in \cref{tab:pooling}. We first perform patch-wise prediction, which is elaborated in \cref{subsec:pooling}. For comparison, we follow WaDIQaM~\cite{bosse2017deep} and IQMA~\cite{guo2021iqma} to use spatial pooling that combines max-pooling and average-pooling in spatial dimension to obtain a score vector $S\in \mathbb{R}^{1\times 1 \times C}$. The final score is the weighted sum of the score vector and the final result is shown in the second row of \cref{subsec:pooling}. Then we try to combine the previous two pooling method and propose to use the average of the output score from patch-wise prediction and spatial pooling in the third experiment. Patch-wise prediction module proposed in AHIQ performs better than the other two, and experimental results further prove the validity of the patch-wise prediction operation. It confirms our previous claim that different regions should contribute differently to the final score.

\subsection{NTIRE 2022 Perceptual IQA Challenge}
This work is proposed to participate in the NTIRE 2022 perceptual image quality assessment challenge~\cite{gu2022ntire}, the objective of which is to propose an algorithm to estimate image quality consistent with human perception. The final results of the challenge in the testing phase are shown in~\cref{tab:final}. Our ensemble approach won the first place in terms of PLCC, SROCC, and main score.
\begin{table}[ht]
\centering
\caption{The results of NTIRE 2022 challenge FR-IQA track on the testing dataset. This table only shows part of the participants and best scores are~\textbf{bolded}.}
\begin{tabular}{cccc}
\toprule[1.2pt]
Method & PLCC & SROCC & Main Score \\ \hline
Ours
& \textbf{0.828} & \textbf{0.822} & \textbf{1.651} \\
\nth{2} 
& 0.827 & 0.815 & 1.642 \\
\nth{3} & 0.823 & 0.817 & 1.64 \\
\nth{4} & 0.775 & 0.766 & 1.541 \\
\nth{5} & 0.772 & 0.765 & 1.538 \\ \toprule[1.2pt]
\end{tabular}
\label{tab:final}
\end{table}

\section{Conclusion}
In this paper, we propose a novel network called Attention-based Hybrid Image Quality Assessment Network (AHIQ), for the full-reference image quality assessment task. The proposed hybrid architecture takes advantage of the global semantic features captured by ViT and local detailed textures from a shallow CNN during feature extraction. To help CNN pay more attention to the salient region in the image, semantic information from ViT is adopted to guide deformable convolution so that model can better mimic how humans perceive image quality. Then we further propose a feature fusion module to combine different features. We also introduce a patch-wise prediction module to replace spatial pooling and preserve information in the spatial dimension. Experiments show that the proposed method not only outperforms the state-of-the-art methods on standard datasets, but also has a strong generalization ability on unseen samples and hard samples, especially GAN-based distortions. The ensembled version of our method ranked first place in the FR track of the NTIRE 2022 Perceptual Image Quality Assessment Challenge.

\vspace{2pt}
\noindent\textbf{Acknowledgment.} This work was supported by the Key Program of the National Natural Science Foundation of China under Grant No. U1903213 and the Shenzhen Key Laboratory of Marine IntelliSense and Computation under Contract ZDSYS20200811142605016.

{\small
\bibliographystyle{ieee_fullname}
\bibliography{egbib}
}

\end{document}